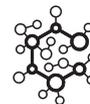

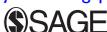

COLLECTIVE INTELLIGENCE

*Research Article*

# Kill chaos with kindness: Agreeableness improves team performance under uncertainty

## Soo Ling Lim and Peter J Bentley
[1]Department of Computer Science, University College London, UK

## Randall S Peterson
[2]London Business School, UK

## Xiaoran Hu
[3]London School of Economics and Political Science, UK

## JoEllyn Prouty McLaren
[4]TalentSage LLC, Minnesota, USA

## Abstract
Teams are central to human accomplishment. Over the past half-century, psychologists have identified the Big-Five cross-culturally valid personality variables: Neuroticism, Extraversion, Openness, Conscientiousness, and Agreeableness. The first four have shown consistent relationships with team performance. Agreeableness (being harmonious, altruistic, humble, and cooperative), however, has demonstrated a non-significant and highly variable relationship with team performance. We resolve this inconsistency through computational modelling. An agent-based model (ABM) is used to predict the effects of personality traits on teamwork, and a genetic algorithm is then used to explore the limits of the ABM in order to discover which traits correlate with best and worst performing teams for a problem with different levels of uncertainty (noise). New dependencies revealed by the exploration are corroborated by analyzing previously unseen data from one of the largest datasets on team performance to date comprising 3698 individuals in 593 teams working on more than 5000 group tasks with and without uncertainty, collected over a 10-year period. Our finding is that the dependency between team performance and Agreeableness is moderated by task uncertainty. Combining evolutionary computation with ABMs in this way provides a new methodology for the scientific investigation of teamwork, making new predictions, and improving our understanding of human behaviors. Our results confirm the potential usefulness of computer modelling for developing theory, as well as shedding light on the future of teams as work environments are becoming increasingly fluid and uncertain.

**CCS CONCEPTS** • Computing methodologies—modeling and simulation—simulation types and techniques—agent/discrete models • Computing methodologies—machine learning—machine learning approaches—bio-inspired approaches—genetic algorithms

**Corresponding author:**
Soo Ling Lim, Department of Computer Science, University College London, Gower Street, London WC1E 6BT, UK.
Email: s.lim@cs.ucl.ac.uk

## Significance Statement

Using a novel interdisciplinary computational modelling approach based on agent-based modelling and genetic algorithms that is grounded in personality psychology, we identify that higher average team agreeableness correlates with better performance for tasks with uncertainty. Agreeableness is a personality trait that corresponds to compliance, trust, tendermindedness, and social harmony. Psychologists have found inconsistent results for this trait in teamwork—some teams benefit from it while others do not. We resolve these conflicting findings and show a new role of agreeableness through the use—for the first time in this field—of an agent-based model to simulate how individuals with different personality traits perform as a team. We confirmed the prediction with data gathered over more than a decade comprising 3698 individuals in 593 teams on tasks with and without uncertainty.

## Introduction

Human behavior was thought by many to be largely unpredictable because of factors such as randomness and individual differences (Cziko, 1989), despite decades of research by psychologists. These ideas now seem dated. With enough data, it is possible to be remarkably accurate making predictions about us, some of which form a fundamental part of our modern world. Using data, we can predict the likes or dislikes of individuals for specific products or online content—a fact exploited by recommender systems worldwide (Resnick and Varian, 1997). Modeling human behavior is also a commonplace in architecture design (e.g., when analyzing access routes to fire exits (2011)) and economics (e.g., when studying trust networks or styles of auction (Smith, 1991)).

However, some human behaviors remain inherently opaque. Gather individuals together as a team and ask them to complete a task—especially if uncertainty is involved—and the likely success of one team compared to another equally qualified team becomes far less clear. Why is it that one team might break down into intense debate or heated arguments that stall progress, while another might gel as a team and excel in finding good solutions rapidly? Any method that might provide some clues to the likely success or failure of teams could have significant impact and help organizations allocate appropriate individuals to form more effective teams.

Task uncertainty adds to the challenge of understanding teams. It is defined as a lack of predictability associated with inputs, processes, and outputs of the broader system within which the work is performed (Cordery et al., 2010). Task uncertainty may thus be differentiated from task complexity (Wood, 1986), in that it is a property of the immediate task environment, rather than the task itself. In optimization, this is equivalent to the distinction between problem noise and difficulty. At the extreme, task uncertainty may include tasks with *deep uncertainty* (Marchau et al., 2019)—tasks with outcomes that may have multiple plausible futures and where it may be hard to determine key factors within models such as interaction between variables or desirability of outcomes; we focus on tasks where such factors can be defined.

To date, most approaches to the problem of predicting the performance of human teams have been largely empirical (Muthukrishna and Henrich, 2019). Statistical studies of data produced by individuals as they attempt different tasks are used to derive statistical models or psychological theories. Some attempts to use computer modelling have been made, but their predictions are largely unverified with data (Ahrndt et al., 2015; Lim and Bentley, 2018, 2019b; Salvit and Sklar, 2012). Neither approach explains teamwork adequately—the former suffers from a lack of data on teams and may not offer insights into why certain combinations of individuals may work better than others, while the latter suffers from a lack of rigorous corroboration.

For the first time, this work combines and enhances both approaches by constructing an agent-based model to explicitly represent the assumptions made by the well-established five-factor model of personality and using a genetic algorithm in order to make predictions of the relative success of different teams when tackling problems of differing uncertainty. The quality of these predictions is then assessed through in-depth analysis of one of the largest team performance datasets comprising 3698 individuals in 593 teams collected over a 10-year period.

This research aims to show that evolutionary computation can explore the extremes of an agent-based model to result in (A1) valid predictions as confirmed by the literature and (A2) novel predictions that can be corroborated by real-world data and used to help explain real phenomena.

In the course of this work, we make the following contributions to the field of evolutionary computation: (i) A novel particle swarm optimization (PSO) algorithm that uses different strategies per particle (agent) to form an agent-based model (ABM) of human behavior. This is the first work in the field that rigorously maps personality traits to agent behavior. (ii) The validation of predictions arising from the use of a genetic algorithm (GA) with the ABM is performed by comparing with the literature (on known trends), and with a previously unseen dataset (to corroborate new findings). Such validation is rarely performed in the literature. (iii) The application of evolutionary computation (GAs, PSO, and ABMs) on a new domain of psychology. (iv) A demonstration of a scientific methodology based on evolutionary computation: using GAs to search the limits of an ABM to suggest hypotheses, which can then be proven by the analysis of real-world data (Bryson et al., 2007).

While these contributions highlight an epistemological approach to explore relationships, this work also establishes clear ontological relationships between agreeableness, task uncertainty, and team performance, making a significant contribution to the field of team psychology. Agreeableness (being harmonious, altruistic, humble, and cooperative) has to date demonstrated a non-significant and highly variable relationship with team performance: some studies report agreeableness to improve team performance (Barrick et al., 1998; Bell, 2007; Neuman et al., 1999; Neuman and Wright, 1999; Peeters et al., 2006), others find no correlation between agreeableness and team performance (Barry and Stewart, 1997; Bell, 2007; Day and Carroll, 2004; Mohammed and Angell, 2003; Peeters et al., 2006). This work resolves this inconsistency: the combination of GA and ABM predicts that team member agreeableness improves team performance in tasks with uncertainty but not in tasks without uncertainty, and we corroborate the prediction with a new analysis of real-word data on teamwork.

The rest of the paper is organized as follows. Section 2 describes background literature, Section 3 describes the method, Section 4 the results, Section 5 the analysis, Section 6 limitations and future work, and Section 7 concludes.

## Background

### Statistical and probabilistic models of human behavior

The increasing prevalence of data has led many researchers to focus on data-driven approaches. Gonzalez et al. (2008) studied the trajectory of mobile phone users and found that human trajectories show a high degree of temporal and spatial regularity. Almeida and Azkune (2018) created a deep learning architecture based on long short-term memory networks that offers a probabilistic model to identify anomalous user behaviors and help detect risks related to mild cognitive impairment and frailty. Sagl et al. (2012) used user-generated data in mobile networks and voluntarily published information on social media platforms to understand how collective social activity shapes urban systems. Fabi et al. (2013) modelled user behavior in the context of real energy use and applied to a case study. The methodology, based on a medium/long-term monitoring, is aimed at shifting toward a probabilistic approach for modelling the human behavior related to the control of indoor environment.

Despite these works on individual behaviors, data on team performance are still almost non-existent, as it is difficult to gather in any meaningful scale or with any useful degree of accuracy.

### Evolving agent-based models

Agent-based models (ABMs) are increasingly used to investigate the processes, mechanisms, and behaviors of complex social systems due to their ability to capture the nonlinear dynamics of social interactions. The interaction between rules and agents can often lead to emergent behavior useful in modelling complex systems such as teams, and ABMs are used in a wide range of research areas including sociology, biology, epidemiology, and ecology (Bentley, 2009; Cuthbert et al., 2017; Farmer and Foley, 2009). Evolutionary computation techniques have been used extensively with ABMs, as they exhibit similar properties to biological systems (e.g., genotype/phenotype mapping, adaptation, and evolution), and they can handle large search spaces efficiently (Morrall, 2003; Whigham and Fogel, 2006). Existing models of nature-inspired evolutionary ecosystems include Echo (Holland, 1992), Swarm (Minar et al., 1996), and SugarScape (Epstein and Axtell, 1996).

There has also been increasing research that uses evolutionary ABMs to explore and understand human interactions (Bentley and Lim, 2022). Nikolic and Dijkema (2010) developed an evolutionary modeling process for creating evolving, complex ABMs to understand the evolution of large-scale sociotechnical systems such as infrastructures for energy, water, and transport. Lim et al. (2015) evolved app developer strategies in the app store to understand the relationship between stability and fitness. Other examples include financial markets (Palmer et al., 1994; Takadama et al., 2001), human resource and organizational culture (Chen et al., 2002), foreign exchange markets (Izumi and Ueda, 2001), labor markets (Tassier and Menczer, 2001), and electricity markets (Bunn and Oliveira, 2001; Nicolaisen et al., 2001).

### Agent-based models of personality in collaboration

Agent-based modelling has a long history of success in many related fields from economics and cooperative behaviors, to social conflict, civil violence, and revolution. However, its use remains very limited in studies of human psychology and their effects on teamwork.

The personality of individuals in a team provides a key role in understanding whether the team will be successful or not (Barrick et al., 1998; Barry and Stewart, 1997; Bell, 2007; Mohammed and Angell, 2003; Moynihan and Peterson, 2001; Neuman and Wright, 1999; Neuman et al., 1999; Peeters et al., 2006; Peterson et al., 2003). Human personality is classically divided into five basic, universal traits known collectively as the Big-Five, namely, (i) neuroticism—a tendency to experience negative emotions such as anxiety and anger, (ii) extraversion—an inclination to be gregarious, assertive, and active, (iii) openness to experience—the propensity to be intellectually curious and prefer variety, (iv) agreeableness—a predisposition toward working collaboratively and building social harmony, and (v) conscientiousness—the propensity for orderliness and self-discipline (Costa and MacCrae, 1992; Goldberg, 1990).

Recently, researchers in agent-based modelling have begun using personality to determine agent behaviors. Early models have interpretations that loosely match psychology literature, are context dependent, and largely unverified (Ahrndt et al., 2015; Salvit and Sklar, 2012). More recently, Lim and Bentley (2018) created a more general, problem-independent computational model aiming to improve our understanding of the effects of personality on teamwork. Initial work used Jung's type theory as the personality model to determine agent behavior (Lim and Bentley, 2018) and examined the effects of team diversity in individual backgrounds (Lim and Bentley, 2019b), the effects of dynamic problems (Lim and Bentley, 2019a), and the effects of reward sensitivity (Guo et al., 2020).

In this work, we created a new model using the Big-Five trait theory as the personality model, following the principles established in previous works (Guo et al., 2020; Lim and Bentley, 2018, 2019a, 2019b). Big-Five is widely used in modern psychology research (Costa and MacCrae, 1992) and regarded as having better functional evidence that the different dimensions of human personality may be associated with specific structural neuroanatomic correlates (Sampaio et al., 2014).

## Method

In order to investigate the two hypotheses, we developed an agent-based model (ABM) combined with a genetic algorithm to investigate the effects of personality on team performance, for tasks with different levels of noise. Our ABM is inspired by particle swarm optimization (PSO) (Kennedy et al., 2001), which simulates social behavior represented by the movement of individual organisms in a swarm (Arrow et al., 2000). In our model, the swarm represents the team, and each agent represents an individual in the team searching for a good solution to their team task, each with their own perceptions, memory, actions, and personalities.

### Agent-based model

We developed an ABM of human collaboration to simulate the differing behaviors of people according to their Big-Five personality. Our ABM has the following key abstractions. (i) Team goal: Inspired by previous works (Lim and Bentley, 2018, 2019b), we model the shared goal of each agent as the group task to optimize a simple two-dimensional parabola function with domain comprising the possible positions of the agent and range corresponding to the solution space. (ii) Agent behavior: Agent behavior is determined by its personality. We model the current mental state of each agent by giving it a position in a two-dimensional solution space (denoting the solution its mind has found so far), a velocity vector (denoting the direction and speed of its thought process), and acceleration vectors (representing the force of ideas and influences that modify the direction and speed of thought, determined by its Big-Five personality traits). (iii) Information exchange: We model information sharing between agents as they work as a team to solve the problem. The exact type of information perceived by each agent and its use is determined by its Big-Five personality traits, for example, extraverted individuals interact with more members of their team than introverted individuals, individuals high on openness explore more diverse solutions than less open individuals, and agreeable individuals take other team members' opinions into account when making decisions. (iv) Task uncertainty: To model tasks without uncertainty, agents perceive the quality of solutions perfectly; to model tasks with uncertainty, agents perceive the quality of solutions with random error, that is, the degree of noise in the problem is increased.

All abstractions were derived from psychology literature and the effects of individual traits validated (described later). No attempt was made to calibrate the model in order to "design in a desired result" for the experiments, and indeed the complexity of the model prohibits such parameter tuning. The dataset used for corroboration of final results was not used for model design or validation.

The algorithm of the model is illustrated in Figure 1 and detailed in the following subsections.

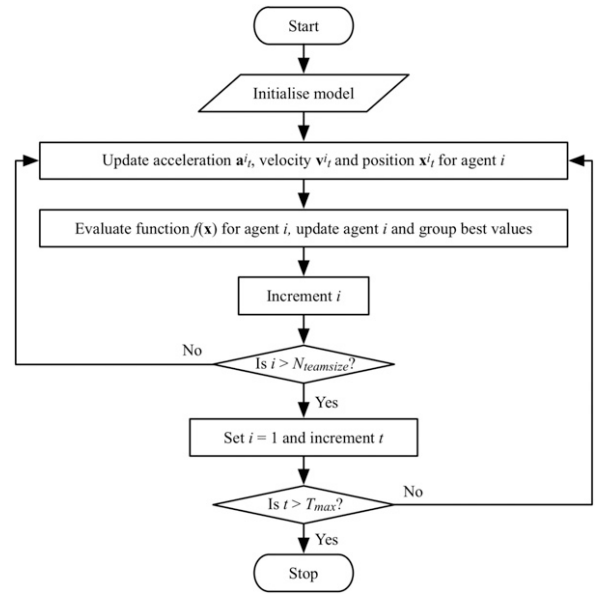

**Figure 1.** Algorithm of the model.

*Initialize.* The model is initialized with a problem space $\mathbf{D} \in \mathbb{R}^n$, an objective function $f(\mathbf{x})$, the number of timesteps $T_{max}$ to run the model, a team of $N_{teamsize}$ agents, and group best $f_{best}^g$ representing the best $f_{best}^i$ and group best position $\mathbf{x}_{best}^g$ representing the corresponding $\mathbf{x}_{best}^i$. Each agent $i \in \{1, \ldots, N_{teamsize}\}$ is initialized with Big-Five personality traits $N^i, E^i, O^i, A^i, C^i$, the value of each trait is a real number $R : Trait_{min} \le R \le Trait_{max}$, a position $\mathbf{x}_0^i \in \mathbf{D} : \mathbf{x}_{min} \le \mathbf{x}_0^i \le \mathbf{x}_{max}$, a random velocity $\mathbf{v}_0^i \in \mathbb{R}^n : -\mathbf{v}_{init} \le \mathbf{v}_0^i \le \mathbf{v}_{init}$, and personal best $f_{best}^i = f(\mathbf{x}_0^i)$ and personal best position $\mathbf{x}_{best}^i = \mathbf{x}_0^i$. We fix $T_{max}$ to represent a fixed time in which the task is to be tackled. Here, we model a common type of task in teamwork called disjunctive tasks where the performance of the overall team is measured by the best individual performer (Steiner, 1972).[1]

*Update.* For each timestep $t \in \{1, \ldots, T_{max}\}$, each agent $i$'s position $\mathbf{x}_t^i$ is updated using (1)

$$\mathbf{x}_t^i = \mathbf{x}_{t-1}^i + \mathbf{v}_t^i \tag{1}$$

Velocity $\mathbf{v}_t^i$ and acceleration $\mathbf{a}_t^i$ are calculated as follows: Acceleration $\mathbf{a}_t^i$ is used to change the direction and speed of thought, as guided by the agent's personality. Interpretations of the literature developed through several iterations of previous empirical studies (Lim and Bentley, 2018, 2019a) were created to represent Big-Five personality traits (openness to experience, conscientiousness, extraversion, agreeableness, and neuroticism) appropriately and were not tuned in order to achieve any specific result in later experiments.

Each trait is modelled as a scale between 0.0 to 1.0, where the likelihood of the behavior corresponding to the trait is proportional to the agent's score in the trait (Fleeson and Gallagher, 2009). For example, if agent $i$'s Extraversion score is 0.8, it means at each timestep $t$, the agent has a probability of 0.8 of being extraverted ($E_t^i$ = True) and 0.2 of not being extraverted ($E_t^i$ = False). The agent's personality at timestep $t$ is thus $N_t^i, E_t^i, O_t^i, A_t^i$, and $C_t^i$. The behavior of agent $i$ based on its personality at timestep $t$ is described as follows and is based on existing Big-Five literature (Costa and McCrae, 1992). Equations are defined in Table 1, and constants are defined in Table 2, and are used according to Algorithm 1.

---

ALGORITHM 1: Agent $i$ behavior at timestep $t$

If $N_t^i$ = True and fitness $f(\mathbf{x}_t^i)$ not improving in $\frac{N_{hist}}{2}$ of $N_{hist}$ previous timesteps, then

    With probability $P_{Neu}$, velocity $\mathbf{v}_t^i$ and acceleration $\mathbf{a}_t^i$ are defined in Eqn (a)

    Otherwise, velocity $\mathbf{v}_t^i$ and acceleration $\mathbf{a}_t^i$ are defined in Eqn (b)

Else if $C_t^i$ = False and with probability $P_{Con}$, then velocity $\mathbf{v}_t^i$ and acceleration $\mathbf{a}_t^i$ are defined in Eqn (c)

Else:

    Set of acceleration vectors $AV = \{\}$

---

$AV \cup \{\mathbf{a}\}$, where $\mathbf{a}$ is defined in Eqn (d) such that if $O_t^i$ = True, $(\mu, \sigma) = (\mu_{large}, \sigma_{large})$ else $(\mu, \sigma) = (\mu_{small}, \sigma_{small})$

If $E_t^i$ = True, then $AV \cup \{\mathbf{a}\}$, where $\mathbf{a}$ is defined in Eqn (e) and $k = N_{neigh\_large}$

Else $AV \cup \{\mathbf{a}\}$, where $\mathbf{a}$ is defined in Eqn (f) and $k = N_{neigh\_small}$

If $A_t^i$ = True, then $AV \cup \{\mathbf{a}\}$, where $\mathbf{a}$ is defined in Eqn (g)

If $C_t^i$ = False, then $AV \cup \{\mathbf{a}\}$, where $\mathbf{a}$ is defined in Eqn (h)

Finally, acceleration $\mathbf{a}_t^i$ is the average of all acceleration vectors in $AV$ and velocity $\mathbf{v}_t^i$ is calculated as $\mathbf{v}_t^i = \mathbf{v}_{t-1}^i + \mathbf{a}_t^i$. If $|\mathbf{v}_t^i| > \mathbf{v}_{max}$, it is scaled to equal $\mathbf{v}_{max}$, in order to prevent excessive speed.

End

---

*Evaluate.* Agent $i$'s fitness at timestep $t$ is evaluated as $f_t^i$ as described in (2)

$$f_t^i = f(\mathbf{x}_t^i) + X_t^i, where\ X_t^i \sim \mathbf{U}(-noise, noise) \tag{2}$$

Finally, the agent's personal best $f_{best}^i$, the agent's personal best position $\mathbf{x}_{best}^i$, group best $f_{best}^g$, and group best position $\mathbf{x}_{best}^g$ are updated.

### Settings and function

The model was initialized with an objective function $f(\mathbf{x})$ as described in (3)

$$f(x, y) = -\sqrt{x^2 + y^2} \tag{3}$$

The function was normalized such that $f(\mathbf{x}) \in [0, 1] : \forall x \in [\mathbf{x}_{min}, \mathbf{x}_{max}]$ and $noise \in \{0.0, 0.1, 0.2\}$, thus $noise = 0.0$ corresponds to 0% noise, $noise = 0.1$ corresponds to 10% noise, and $noise = 0.2$ corresponds to 20% noise. Figure 2 shows the heat map and surface plot. The function represents a simple problem with a smooth gradient ($noise = 0\%$), such as tasks with definitive solutions, such as accounting. When simulating noisy problems ($noise = 10\%$ or 20%), such as open-ended tasks, such as entrepreneurship and managing organizational behavior.

### Model validation

In order to validate the accuracy of predictions made by the model and ensure its behaviors are representative of reality, a series of preliminary validations were performed, following principles of good modelling practice (Bentley, 2009). These involved running the simulation with the settings in Table 2. $N_{teamsize}$ was set to be 6. We performed a parameter sweep, investigating the full range of a trait from 0.0 to 1.0, incrementing by 0.1 (a grid search of trait values). For each increment, the model was run 100 times, the other traits were randomized, and average team performance was recorded. This was repeated for *noise* = 0%, 10%, and 20%. The trends in average team performance were then compared with existing literature on teamwork.

**Table 1.** Model equations and implementation.

| Eq | Implementation | Interpretation | Justification |
|---|---|---|---|
| (a) | $\mathbf{v}_t^i = 0, \mathbf{a}_t^i = 0$ | Stop moving | Simulating discouragement and withdrawal for Neuroticism. Neuroticism is associated with the feeling of discouragement and withdrawal when experiencing setbacks (Costa and MacCrae, 1992; DeYoung et al., 2007; Komarraju and Karau, 2005) |
| (b) | $\mathbf{v}_t^i = 0, \mathbf{a}_t^i = \mathbf{a}_{random} \in \mathbb{R}^n : -\mathbf{a}_{large} \leq \mathbf{a}_{random} \leq \mathbf{a}_{large}$ | Accelerate at a random speed toward a random direction | Simulating impulsiveness and volatility for Neuroticism. Neuroticism is associated with impulsiveness and immoderation when experiencing setbacks (Costa and MacCrae, 1992; DeYoung et al., 2007) |
| (c) | $\mathbf{v}_t^i = 0, \mathbf{a}_t^i = 0$ | Stop moving | Simulating procrastination from lack of Conscientiousness. Lack of conscientiousness is associated with lack of self-discipline, not dutiful, and lack of dependability (Costa and MacCrae, 1992; DeYoung et al., 2007) |
| (d) | $\sum_{j=1}^{3} r_j(\mathbf{c}_j - \mathbf{x}_{t-1}^i)$ where $\mathbf{c}_1, \mathbf{c}_2,$ and $\mathbf{c}_3$ are the top 3 candidates derived $\mathbb{C}_t^i$ sorted in the order of decreasing $f(\mathbf{x})$ with $f(\mathbf{c}_1) \geq f(\mathbf{c}_2) \geq f(\mathbf{c}_3)$. $\mathbb{C}_t^i = \{\mathbf{x}_0^i, ..., \mathbf{x}_{t-1}^i\} \cup \mathbf{P}$, where $\mathbf{P}$ is the set of points near to $\mathbf{x}_{t-1}^i$. Given $\mathbf{x}_{t-1}^i = (x_1, x_2, ..., x_n)$, $\mathbf{P} = \{(x_1 + \delta, x_2, ..., x_n), (x_1 - \delta, x_2, ..., x_n), (x_1, x_2 + \delta, ..., x_n), (x_1, x_2 - \delta, ..., x_n), ..., (x_1, x_2, ..., x_n + \delta), (x_1, x_2, ..., x_n - \delta)\}$ where $\delta$ is a random number from a normal distribution $N(\mu, \sigma)$ | Evaluate previous positions and new areas where the size of coverage is determined by $N(\mu, \sigma)$ and move toward the top 3 best positions | Simulating intellectual curiosity, where Openness corresponds to higher levels of curiosity (a larger area of exploration) and lack of Openness corresponds to lower levels of curiosity (a smaller area of exploration) (Costa and MacCrae, 1992; DeYoung et al., 2007) |
| (e) | $\mathbf{C}_{t-1}^i - \mathbf{x}_{t-1}^i$ where $\mathbf{C}_{t-1}^i$ is the centroid (arithmetic mean position) of agent $i$'s $k$ nearest neighbors' positions in the previous timestep. If the neighbor is currently extraverted, count its position twice | Move toward neighbors' current positions. If neighbor is currently extraverted, it is more assertive so its position is counted twice | Simulating sociability and assertiveness for Extraversion. Extraversion is associated with high influence and tendency to assert its opinions on others (Costa and MacCrae, 1992; DeYoung et al., 2007) |
| (f) | $\mathbf{x}_{best-t-1}^i - \mathbf{x}_{t-1}^i$ where $\mathbf{x}_{best-t-1}^i$ is agent $i$'s personal best position in the previous timestep and $\mathbf{x}_{t-1}^i$ is the agent's position in the previous timestep | Move toward their own previous best position | Simulating reclusion from lack of Extraversion (Costa and MacCrae, 1992; DeYoung et al., 2007; Komarraju and Karau, 2005) |
| (g) | $\mathbf{C}_{t+2}^i - \mathbf{x}_{t-1}^i$ where $\mathbf{C}_{t+2}^i$ is the centroid (arithmetic mean position) of agent $i$'s $k$ nearest neighbors' positions in the next two timesteps. If the neighbor is currently extraverted, count its position twice. Each agent $i$'s position in the next two timesteps is calculated as $\mathbf{x}_{t+2}^i = \mathbf{x}_{t-1}^i + \mathbf{v}_{t-1}^i + \mathbf{v}_{t-1}^i$ | Get influenced by where neighbors are heading. If neighbor is currently extraverted, it is more assertive, so its position is counted twice | Simulating social harmony for Agreeableness, a tendency to be compassionate and cooperative rather than suspicious and antagonistic toward others. Agreeable individuals value getting along with others. They are generally considerate, kind, generous, trusting and trustworthy, helpful, and willing to compromise their interests with others (Costa and MacCrae, 1992; DeYoung et al., 2007) |
| (h) | $\mathbf{a}_{random} \in \mathbb{R}^n : -\mathbf{a}_{small} \leq \mathbf{a}_{random} \leq \mathbf{a}_{small}$ | Margin of error in acceleration vector | Simulating spontaneity and lack of reliability from lack of Conscientiousness. Low conscientiousness is associated with flexibility and spontaneity, which also appear as sloppiness and lack of reliability (Costa and MacCrae, 1992; DeYoung et al., 2007) |

Figure 3 shows the results of the validation, which demonstrated that the model trends match published trends. We describe the published trends and corresponding model trends as follows:

1. Team performance deteriorates as uncertainty increases (Cordery et al., 2010). In Figure 3, as *noise* increases from 0% (left column) to 20% (right column), average team performance decreases for each trait.
2. Variability in performance increases as uncertainty increases (Cordery et al., 2010). In Figure 3, as *noise* increases from 0% (left column) to 20% (right column), the standard deviation increases for each trait.
3. Neuroticism decreases team performance, especially when levels are very high (Peeters et al., 2006). Figure 3 Row 1 shows that when *noise* = 0%, performance decreases as neuroticism increases, with maximum neuroticism resulting in very poor performance with high variability. This trend is less clear at other noise levels.
4. Extraversion improves team performance (Barrick et al., 1998; Bell, 2007; Peeters et al., 2006), but high levels of extraversion in simple tasks decrease team performance (Moynihan and Peterson, 2001; Peeters et al., 2006). Figure 3 Row 2 shows that when *noise* = 0% (left column), team performance decreases as extraversion increases. When *noise* = 10% and 20% (middle and right columns), team performance increases as extraversion increases.
5. Openness to experience increases team performance in challenging tasks (Bell, 2007; Peterson et al., 2003) but less so in simple tasks (Peeters et al., 2006). Figure 3 Row 3 shows that when *noise* = 10% and 20% (middle and right columns), team performance increases as openness to experience increases. When *noise* = 0% (left column), team performance is similar regardless of the increase.
6. Literature findings for agreeableness have been inconsistent, with some finding positive relationships between agreeableness and team performance (Barrick et al., 1998; Bell, 2007; Neuman et al., 1999; Neuman and Wright, 1999; Peeters et al., 2006), others finding no relationships between agreeableness and team performance (Barry and Stewart, 1997; Bell, 2007; Day and Carroll, 2004; Mohammed and Angell, 2003; Peeters et al., 2006). Figure 3 Row 4 shows that when *noise* = 0%, team performance is similar regardless of increase in agreeableness, but when *noise* = 10% and 20% (middle and right columns), team performance increases as agreeableness increases.
7. Conscientiousness increases team performance and also decreases variability (Barrick et al., 1998; Bell, 2007; Neuman and Wright, 1999; Peeters et al., 2006). Figure 3 Row 5 shows that in all 3 noise levels, higher conscientiousness results in higher performance and smaller standard deviation.

## Using a genetic algorithm to explore the limits of the ABM

Having established that the ABM performs as expected by the literature when examining each personality trait independently, we now revisit the aims of this work, which are to show that evolutionary computation can explore the extremes of an agent-based model to result in (A1) valid predictions as confirmed by the literature and (A2) novel predictions that can be corroborated by real-world data and used to help explain real phenomena. The validation in the previous section shows that each personality trait, when considered independently, appears to correlate with behavioral trends reported in the literature. However, in reality, every member of a team has specific levels of each trait that may interact and alter the way each member interacts in their team. Understanding this complex set of interactions requires a different approach.

The agent-based model is based on PSO, with every agent attempting to find the optimal solution according to its own personality. It follows that for a given problem type, there will be extreme team combinations: specific combinations of agents will perform better on average, other combinations will perform worse. We therefore need a new experimental tool to use in order to examine the limits of the ABM model. When forced to behave as effectively as it is capable of—or as ineffectively—what combinations of agent traits are best? We use a genetic algorithm as our experimental tool for this purpose.

The genetic algorithm (GA) was used to evolve an optimal (and least optimal) combination of agent personalities within the ABM for solving noisy and non-noisy problems (i.e., problems with and without uncertainty, respectively) and in this way predict which teams in the ABM are likely to work more effectively (and less effectively) for different problems. GAs are known to be

**Table 2.** Model constants and values.

| Constants | Values | Explanation |
|---|---|---|
| $T_{max}$ | 100 | The duration for agents to solve the problem |
| $N_{hist}$ | 6 | Neuroticism is associated with low tolerance for stress and the tendency to view minor frustrations as hopelessly difficult, so fitness not improving in 3 out of 6 timesteps is intended to model minor frustrations |
| $P_{Neu}$ | 0.5 | 50:50 chance of impulsiveness versus withdrawal |
| $P_{Con}$ | 0.5 | 50% chance of procrastination |
| $\mathbf{v}_{max}$ | 5.0 | Velocity cap. An individual with excessive velocity would literally become to "set in their ways" and would find it impossible to change its direction of thought in a useful direction |
| $\mathbf{x}_{min}$ | $(-100.0, -100.0)$ | The minimum boundary of the problem space |
| $\mathbf{x}_{max}$ | (100.0, 100.0) | The maximum boundary of the problem space |
| $\mathbf{v}_{init}$ | (1.0, 1.0) | A small initial velocity |
| $(\mu_{large}, \sigma_{large})$ | (12.0, 5.0) | Openness is associated with appreciation for adventure, unusual ideas, curiosity, and variety of experience, so this simulates the exploration of a large area of possible solutions |
| $(\mu_{small}, \sigma_{small})$ | (1.0, 0.01) | Lack of openness is associated with close mindedness and lack of adventurousness, so this simulates the exploration of a small area of possible solutions |
| $r_1, r_2, r_3$ | 0.5, 0.3, 0.2 | Larger weights provide priority to the better solutions |
| $\mathbf{a}_{large}$ | (20.0, 20.0) | Large values were chosen (20% of the problem space size) so that agents can make large movements to simulate impulsiveness |
| $\mathbf{a}_{small}$ | (5.0, 5.0) | Small values were chosen (5% of the problem space size) to simulate unreliability due to lack of conscientiousness |
| $N_{neigh\_large}$ | 5 | Pronounced engagement with the external world and social involvement |
| $N_{neigh\_small}$ | 2 | Lack of engagement and social involvement |

successful in optimizing noisy and discontinuous problems; the problem of finding optimal team members for the ABM is difficult as any small modification made to a personality trait may result in significant changes to interactions in a team, and because the model is probabilistic, results may vary across runs. For a team of $N_{teamsize} = 4$ members as used in the experiments (chosen to ensure computation tractability), their personalities are specified by a real-coded genotype consisting of 20 randomly generated real numbers between $Trait_{min}$ and $Trait_{max}$. For example, the genotype

0.2, 0.3, 0.5, 0.8, 0.4, 0.1, 0.4, 0.6, 0.4, 0.7, 0.8, 0.2, 0.4, 0.7, 0.6, 0.3, 0.3, 0.5, 0.3, 0.2

corresponds to the phenotype

Agent 1 = {N = 0.2, E = 0.3, O = 0.5, A = 0.8, C = 0.4}
Agent 2 = {N = 0.1, E = 0.4, O = 0.6, A = 0.4, C = 0.7}
Agent 3 = {N = 0.8, E = 0.2, O = 0.4, A = 0.7, C = 0.6}
Agent 4 = {N = 0.3, E = 0.3, O = 0.5, A = 0.3, C = 0.2}

To find best ABM teams, a standard canonical GA is used with a population size of $N_{pop}$, each individual solution representing the personalities of a team. For each member of the population, fitness is calculated by decoding the genotype to produce Big-Five personalities for the team members. The agent-based model is run 100 times for the team to produce average team performance. The group best at the end of each run is recorded, and team performance is measured by their average group best, which is the total group best for all runs divided by the total number of runs. Based on fitness, $N_{parents}$ individuals with the highest average group best are chosen as parents. $N_{pop}$ child teams are created using single-point crossover from the parents. One out of the 20 real numbers is selected for mutation. A simple creep mutation of 0.05 is used, capped at $Trait_{min}$ and $Trait_{max}$. The GA is run for $N_{gen}$ generations. GA parameter values were found following preliminary parameter sweep experiments to determine fastest and most effective settings to ensure efficient convergence to optimal solutions and avoid overfitting.

To find worst ABM teams, instead of taking $N_{parents}$ individuals with the highest average group best, we take $N_{parents}$ individuals with the lowest average group best.

We ran the GA 60 times to find 60 best ABM teams for *noise* = 0%, 60 times to find 60 worst ABM teams for *noise* = 0%, 60 times to find 60 best ABM teams for *noise* = 20%, and 60 times to find 60 worst ABM teams for *noise* = 20%. This was to determine whether multiple optima (i.e., more than one combination of top-performing personalities) exist and to ensure reliability of predicted results.

We define the general population to be a typical group composed by randomly selecting individuals from normal human population, which removes extremes such as every individual in the model exhibiting a certain trait all of the time or none of the time. Settings to simulate the general population are as follows

$$Trait_{min} = 0.1 \text{ (for all traits)}$$

$$Trait_{max} = 0.9 \text{ (for all traits)}$$

$$N_{parents} = 5, N_{pop} = 30, N_{gen} = 100$$

The experiments were run on Amazon EC2 Compute Optimized c5.large instances (2 vCPU 4 GiB memory), with Ubuntu Server 16.04 LTS (HVM), SSD Volume Type (C5 instances are designed for running advanced compute-intensive workloads including scientific modelling, distributed analytics, and machine learning inference). The code was written in Python and available to be downloaded and run.[2] There was a total of 240 evolutionary runs, with a total computation time of 318 h. Each run took an average of 1.33 h ($SD = 0.29$). The fastest run took 49.83 min, and the slowest run took 1.74 h.

We compared the averages between best and worst ABM teams and used *t*-test to see if the differences between the means are significant.

## Team performance data

We use a previously unseen real-world dataset of team performance to assess the validity of predictions made by the model. As with all such real-world data, the signal-to-noise ratio is low and the number of features is high, meaning that it is infeasible to apply statistical methods checking for correlations and dependencies without first having some idea of what to look for. The predictions of the ABM provide such cues.

We collected one of the largest team performance data to date. Participants comprised London Business School MBA (Masters of Business Administration) students across 10 academic years, from 2007 to 2016. A total of 3698 participants were involved in the study (male = 2623 [71%], female = 1075 [29%]), with an average age of 28.5 ($SD = 2.5$). Participants were assigned to 593 groups, and performance data consist of average group project scores for multiple courses. Group performance was assessed using group task scores for courses that exist over all 10 years, and assignments remained consistent. A summary of the courses and their group tasks can be found in Supplemental Materials Table 1.

The data were separated into tasks with and without uncertainty (i.e., problems with and without noise) using the criteria outlined by Laughlin and Ellis (Laughlin and Ellis, 1986) where tasks with uncertainty correspond to courses where there is faculty judgment in the assessment of correctness and those with multiple paradigms such as tasks involving the establishment of a new business strategy (i.e., Management, Strategy, Operations, Entrepreneurship, and Marketing) and tasks without uncertainty correspond to courses where there is a shared conceptual system for evaluation of quality (e.g., quantitative problem sets to work through), such as tasks involving mathematics and predefined rules (i.e., Accounting, Economics, and Finance).

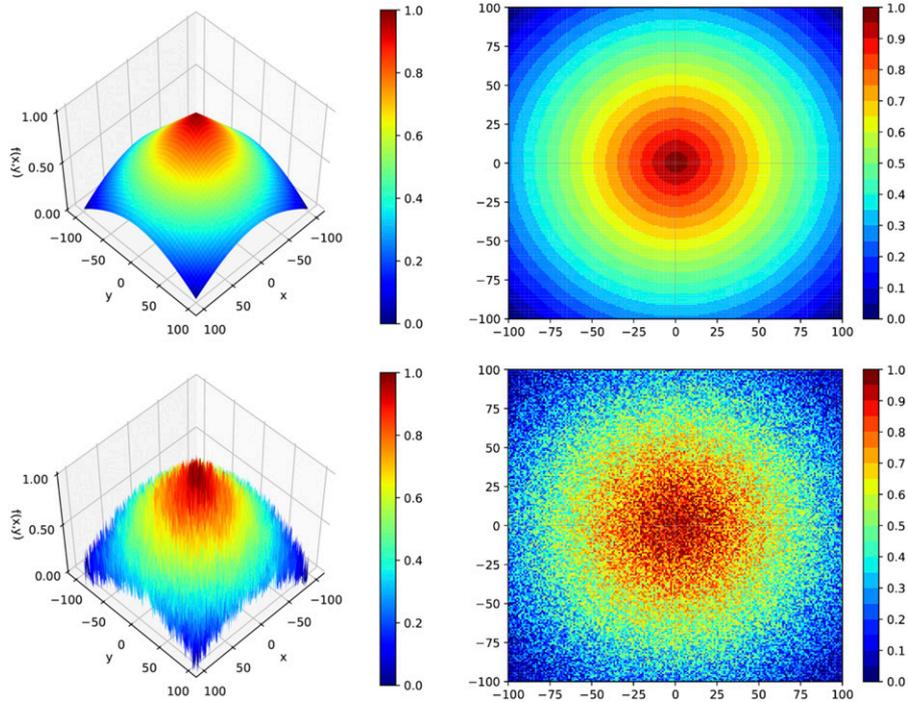

**Figure 2.** Visualization of task used by the agent-based model to represent a simple team task, with solution found in the center of the solution space; *noise* = 0% (top); *noise* = 20% (bottom). Surface plot (left) and heat map (right) for normalized equation (2) with a maximum in (0, 0). Color ranges from blue (minimum) to red (maximum).

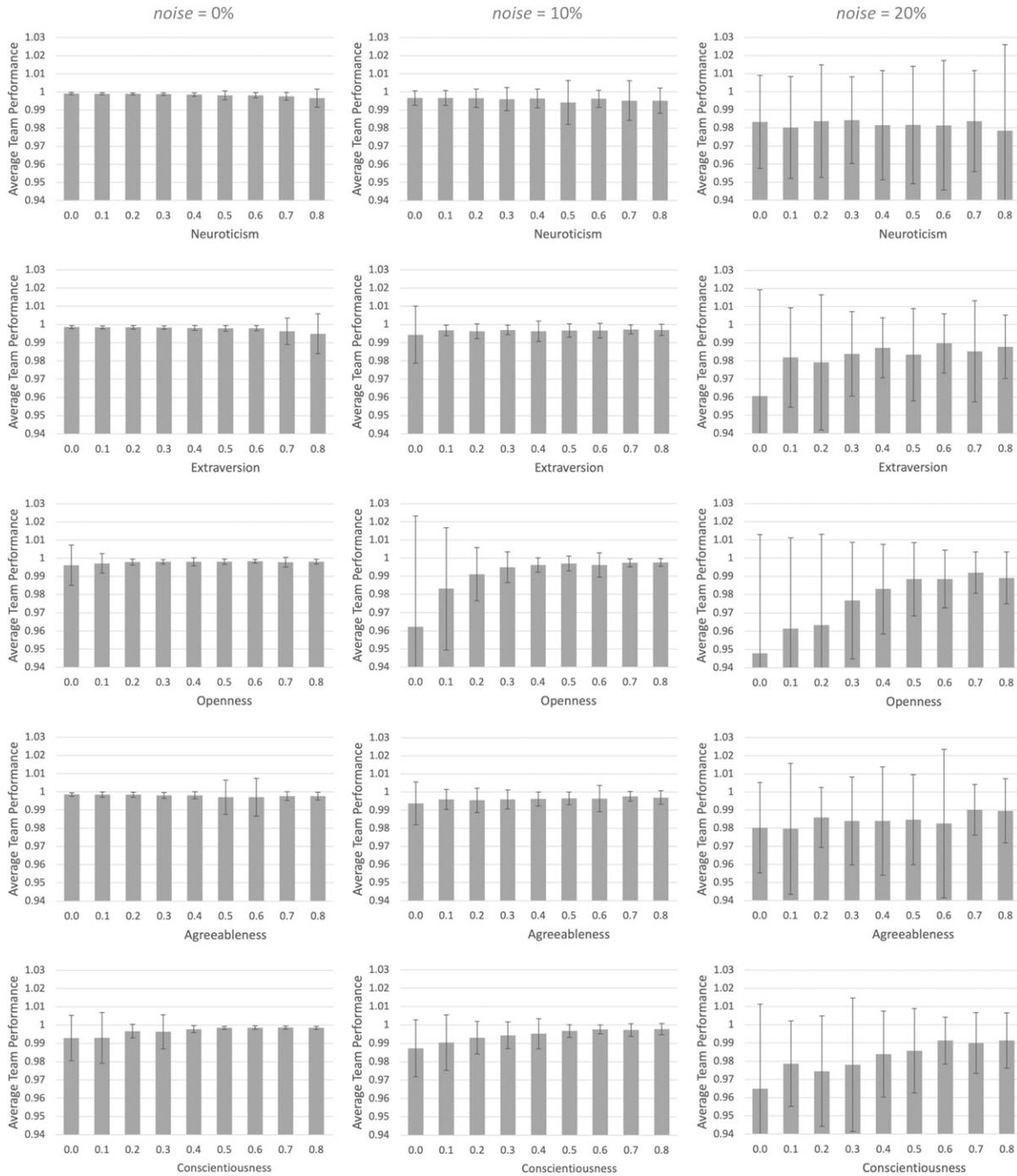

**Figure 3.** Model validation results plotting team performance averaged across 100 runs against teams with mean levels of each trait from 0.0 to 1.0 (with the other traits randomized) for *noise* = 0%, 10%, and 20%. Error bars: mean ± SD.

Average *z*-scores were used to determine group performances. For each course, we calculated the *z*-score per stream per year (as each stream can have different instructors and marked out of different totals), so that they are comparable across years and streams. Each group completed all group projects in their original assigned groups. The group performance is the average *z*-scores for tasks with uncertainty (from the courses Strategy, Entrepreneurship, Managing Organizational Behavior, Operations Management, and Marketing) and tasks without uncertainty (from the courses Management Accounting, Managerial Economics, Financial Accounting, and Finance). We selected the top 60 teams (~10% of total teams) over the 10 years as the best teams and bottom 60 teams (~10% of total teams) over the 10 years as the worst teams. We do so for tasks with and without uncertainty.

Group personality is calculated as the average of individual personalities following (Woolley et al., 2010) and described as follows. Individual personality was assessed using the NEO-PI-R personality inventory (Costa and MacCrae, 1992) (one of the most widely used instruments for measuring Big-Five in psychology research and practice). This test measures the five primary domains of adult personality: Extraversion, Agreeableness, Conscientiousness, Openness to Experience, and Neuroticism. A total of 240 items were answered on a 5-point Likert-scale ranging from strongly disagree to strongly agree. The mean for each scale was calculated by summing up the six facets that compose each five factor for each individual. Scores on each domain were averaged for the group as a whole.[3]

We compared the average personality traits between the top and bottom 60 teams and used *t*-tests to see if the differences between the means are significant. To test the reliability of our results, we also performed an ordinary least squares (OLS) regression analysis to test the relationship between team agreeableness and performance on all 593 teams in our sample. We also performed binary logistic regression analysis to test the relationship between the best and worst teams. We controlled for group size, average Graduate Management Admission Test (GMAT) score, and average years of work experience. The descriptive statistics can be found in Table 3 (tasks without uncertainty) and Table 4 (tasks with uncertainty).

## Results

### A1. Evolutionary computation can explore the extremes of an agent-based model to result in valid predictions as confirmed by the literature

Figure 4 shows the average fitness of teams per generation. It shows that the GA was able to explore the extremes of the ABM successfully. The ABM is behaving as an optimizer (each team tries to find the optimal solution by moving and communicating according to their unique combination of traits); when the GA is used to find the ABM team with the best traits, only small improvements to the best team performance can be achieved, as performance was already good on average. When the GA is used to find the worst performing ABM teams (making the optimization worse), it fares much better, achieving large decreases in performance. The evolved traits for best and worst teams comprise our predictions.

The scatterplots in Figure 5 show the relationship between team performance and the Big-Five traits of each individual team member. As expected, the ABM performs worse for noisy problems, resulting in a clear separation visible in performance on the *y*-axis. Also visible is the convergence (left or right on the *x*-axis) caused by the GA optimizing traits toward specific values.

Figure 6 shows all traits together to enable visual comparison; and the *t*-test results comparing traits and noise levels can be found in Supplemental Materials Table 2. The trends are consistent with the literature. For example, for tasks with uncertainty, best teams have higher extraversion than worst teams because more interaction between team members can make team members more aware of different views, and best teams also have higher openness to experience than worst teams because the tendency to explore more different solutions will produce better results in uncertain environments. In tasks without uncertainty, best teams have lower extraversion than best teams in tasks with uncertainty because interaction is less important. Worst teams have very low conscientiousness because being consistent is a good way to perform well when the task has no uncertainty.

In psychology literature, extraversion, openness, and conscientiousness consistently positively predict team performance (Barrick et al., 1998; Neuman and Wright, 1999) (a finding mirrored in the ABM). Neuroticism consistently negatively predicts team performance (Barrick et al., 1998; Neuman and Wright, 1999), also mirrored in the model. Thus, the predictions made by the ABM are validated by the literature.

### A2. Evolutionary computation can explore the extremes of an agent-based model to result in novel predictions that can be corroborated by real-world data and used to help explain real phenomena

While many of the predictions made by the model are well-known in psychology, one result is novel: the combined effects of noise and agreeableness. Intriguingly, findings from psychology literature for agreeableness have been inconsistent, with field studies generally predicting positive relationships (Barrick et al., 1998; Bell, 2007; Neuman et al., 1999; Neuman and Wright, 1999; Peeters et al., 2006) (except for two field studies, one involving profitability of pizza franchisees (Hofmann and Jones, 2005) and another involving

**Table 3.** Tasks without uncertainty: Descriptive statistics and correlations among study variables.

| Variable | Mean | SD | Min | Max | 1 | 2 | 3 | 4 | 5 | 6 | 7 | 8 |
|---|---|---|---|---|---|---|---|---|---|---|---|---|
| 1. Team performance | 0.00 | 0.66 | −2.18 | 1.51 | | | | | | | | |
| 2. GMAT | 694.31 | 9.95 | 651.67 | 722.33 | .03 | | | | | | | |
| 3. Work experience | 5.51 | 0.67 | 3.67 | 9.57 | .01 | −.09* | | | | | | |
| 4. Team size | 6.07 | 0.56 | 4.50[a] | 8.00 | .09* | −.26** | .12** | | | | | |
| 5. Neuroticism | 1.91 | 0.29 | 1.00 | 2.73 | .01 | −.04 | −.13** | −.01 | | | | |
| 6. Extraversion | 2.60 | 0.28 | 1.67 | 3.37 | .01 | −.06 | −.01 | .02 | −.21** | | | |
| 7. Openness | 2.41 | 0.27 | 1.53 | 3.17 | −.07 | −.11** | .02 | .10* | .04 | .28** | | |
| 8. Agreeableness | 1.75 | 0.26 | 0.95 | 2.67 | .01 | .00 | .01 | .02 | −.13** | .08* | .15** | |
| 9. Conscientiousness | 3.06 | 0.35 | 2.00 | 4.00 | .08* | .13** | .02 | −.07 | −.31** | .17** | −.05 | .07 |

*Note.* $N = 593$ teams; * significant at $p < .05$, two-tailed; ** significant at $p < .01$, two-tailed.
[a]In some teams, a student did not do one of the courses, so the minimum group size, which is the average group size for a specific team, is not a whole number.

**Table 4.** Tasks with uncertainty: Descriptive statistics and correlations among study variables.

| Variable | Mean | SD | Min | Max | 1 | 2 | 3 | 4 | 5 | 6 | 7 | 8 |
|---|---|---|---|---|---|---|---|---|---|---|---|---|
| 1. Team performance | 0.00 | 0.50 | −1.54 | 1.42 | | | | | | | | |
| 2. GMAT | 694.59 | 9.79 | 651.67 | 723.33 | .09* | | | | | | | |
| 3. Work experience | 5.52 | 0.66 | 3.67 | 9.57 | .01 | −.09* | | | | | | |
| 4. Team size | 6.22 | 0.53 | 4.60[a] | 8.00 | .04 | −.23** | .09** | | | | | |
| 5. Neuroticism | 1.91 | 0.27 | 1.00 | 2.71 | −.02 | −.04 | −.12** | .00 | | | | |
| 6. Extraversion | 2.60 | 0.27 | 1.67 | 3.33 | .02 | −.07 | −.01 | .02 | −.22** | | | |
| 7. Openness | 2.40 | 0.27 | 1.50 | 3.17 | .02 | −.12** | .02 | .11** | .04 | .27** | | |
| 8. Agreeableness | 1.75 | 0.26 | 1.00 | 2.67 | .16** | .00 | .02 | −.02 | −.14** | .09* | .15** | |
| 9. Conscientiousness | 3.06 | 0.34 | 2.00 | 4.00 | .06 | .13** | .02 | −.02 | −.31** | .16** | −.05 | .07 |

*Note.* $N = 593$ teams; * significant at $p < .05$, two-tailed; ** significant at $p < .01$, two-tailed.
[a]In some teams, a student did not do one of the courses, so the minimum group size, which is the average group size for a specific team, is not a whole number.

drilling crews (Van Vianen and De Dreu, 2001)), and lab studies generally producing no relationships (Barry and Stewart, 1997; Bell, 2007; Day and Carroll, 2004; Mohammed and Angell, 2003; Peeters et al., 2006) (except for a lab study where student groups worked on research projects (Van Vianen and De Dreu, 2001), and a written assignment for a business project (Bradley et al., 2013)). These inconsistencies imply that there are moderators affecting agreeableness and team performance, but it was unclear what they were.

Our result demonstrates for the first time that task uncertainty is a primary contingency in the relationship between agreeableness and team performance.

In more detail, the model predicted that agreeableness positively predicts group performance for tasks with uncertainty (Figures 5 and 6). Specifically, the model predicted (i) significant higher agreeableness for best teams than worst teams for tasks with uncertainty [$t(118) = 15.60$, $p < .001; t(118) = 24.28, p < .001$]; (ii) significant lower agreeableness for best teams than worst teams for tasks without uncertainty [$t(118) = 38.88, p < .001; t(68.18) = 46.90, p < .001$]; (iv) significant lower agreeableness for best teams than worst teams for tasks without uncertainty [$t(83.56) = −27.77, p < .001; t(85.39) = −121.16, p < .001$]. Supplemental Materials Table 2 shows the $t$-test results for all traits.

In order to establish whether this prediction has merit, the previously unseen team performance dataset was analyzed in detail. The results of this analysis confirm the same effects as predicted by the model. For tasks with uncertainty, best teams have higher agreeableness than worst teams [$t(118) = 2.92, p < .01$], and worst teams in tasks with agreeableness for worst teams working on tasks with uncertainty than worst teams working on tasks without uncertainty [$t(105.26) = −11.72, p < .001; t(95.49) = −55.35, p < .001$]; (iii) significant higher agreeableness for best teams working on tasks with uncertainty than best teams working on tasks without uncertainty [$t(106.62) = 38.88, p < .001; t(68.18) = 46.90, p < .001$]; (iv) significant lower agreeableness for best teams than worst teams for tasks without uncertainty [$t(83.56) = −27.77, p < .001; t(85.39) = −121.16, p < .001$]. Supplemental Materials Table 2 shows the $t$-test results for all traits.

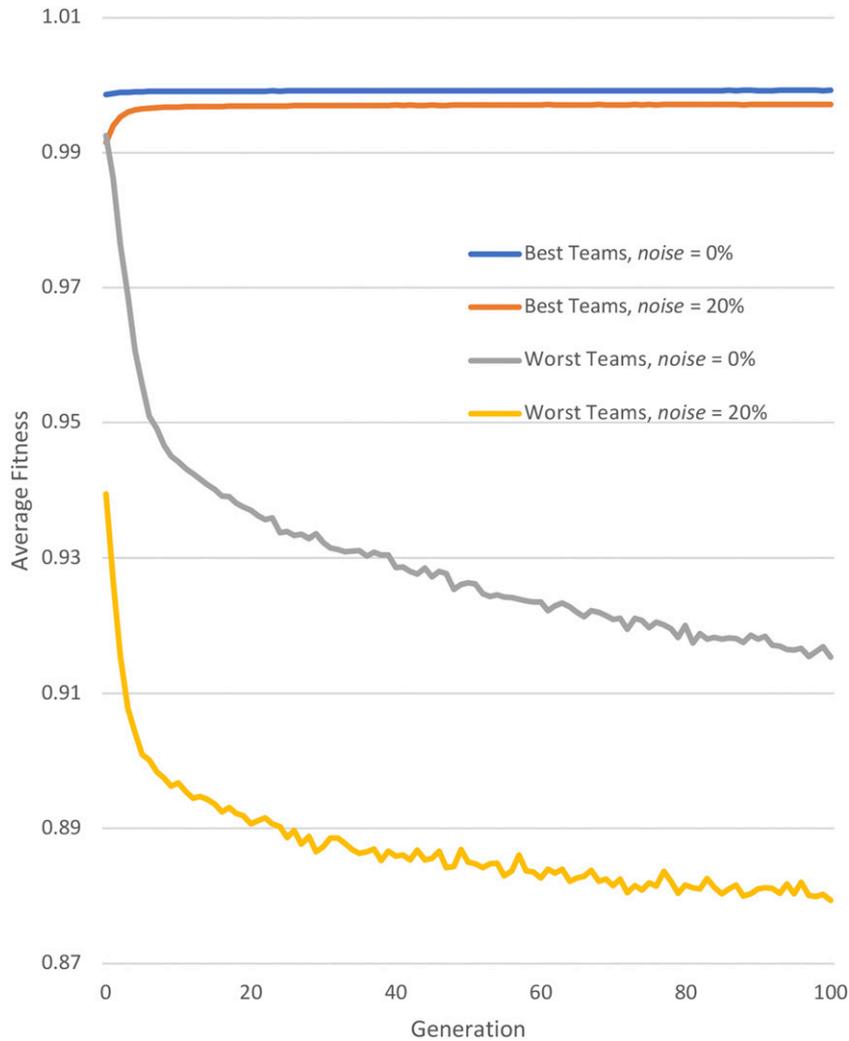

**Figure 4.** Average fitness of best and worst teams across 60 runs over generations. Plots start at first generation.

uncertainty have lower agreeableness than worst teams in tasks without uncertainty $[t(118) = -3.65, p < .001]$. The differences in the modelled comparisons are greater than observed in the data, a finding anticipated due to the considerable noise within such real-world data compared to the clean environment of the ABM (Kim et al., 2017). We also found that best teams in tasks with uncertainty have higher openness than best teams in tasks without uncertainty $[t(118) = 2.32, p < .05]$ (supported by the model), and in tasks without uncertainty, best teams have lower openness than worst teams $[t(118) = -2.68, p < .01]$ (not supported by model). Supplemental Materials Table 3 shows the *t*-test results for all traits.

The results for ordinary least squares (OLS) regression analysis (Table 5) demonstrate that agreeableness positively predicts team performance only for tasks with uncertainty, where agreeableness $(\beta = 0.31, p < .01)$ and GMAT $(\beta = 0.01, p < .05)$ were significant predictors. For tasks without uncertainty, performance was predicted by conscientiousness $(\beta = 0.18, p < .05)$ and team size $(\beta = 0.13, p < .01)$. The findings remain largely unchanged with or without controls and are consistent with existing findings (Bell, 2007), adding confidence to the results of the current study. Binary logistic regression with the same controls to predict best and worst teams were also produced consistent results (Table 6). For tasks with uncertainty, agreeableness and GMAT were significant predictors of performance; for tasks without uncertainty, openness and team size were significant predictors of performance.

The predictions made by the model assumed a general population, but the dataset comprised a very specific sample of individuals—London Business School MBA students—who have a range of traits that differ in important ways. For example, the individuals in the dataset were on average more conscientious and showed less variability in their other traits than the general population. In order to check if the model still produced the same predictions when

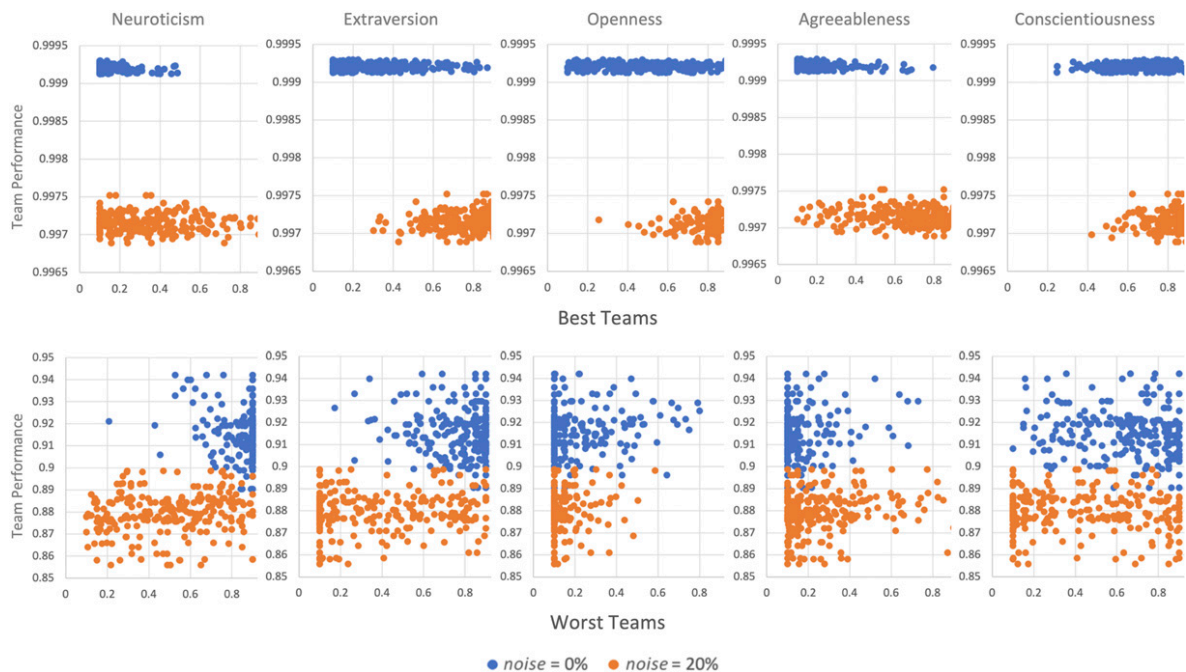

**Figure 5.** Scatter plots showing the relationship between team performance and the Big-Five traits of each team member for the 60 optimized teams. For clarity, we use different *y*-axis scales for Best Teams (top row) and Worst Teams (bottom row).

calibrated to match the sample data, the system was run again, with the following settings

$$Trait_{min} = \{N=0.25, E=0.42, O=0.38, A=0.25, C=0.50\}$$

$$Trait_{max} = \{N=0.68, E=0.83, O=0.79, A=0.67, C=1.00\}$$

$$N_{parents} = 20, N_{pop} = 50, N_{gen} = 100$$

There were a total of 240 simulations, with a total computation time of 584 h. Each simulation took an average of 2.43 h (SD = 0.34). The fastest simulation took 1.9 h, and the slowest simulation took 2.97 h (almost double the time required previously as it was necessary to increase $N_{pop}$ and $N_{parents}$ in order to enable the GA to optimize this more constrained problem).

The predictions made by this more constrained model were consistent with the original model. Similar differences in Figure 7 can be observed, caused by the smaller ranges of traits in the sample population. The prediction for agreeableness remained the same (see Figure 8).

These results provide substantial and novel evidence that task uncertainty moderates when agreeableness is predictive of team performance, demonstrating the context dependence of the effects of agreeableness. The results also provide an explanation for the inconsistency and nonreplicability of past research findings—reconciling the findings that most studies that reported positive effects of agreeableness were conducted in the noisy environment of the field, while most studies that reported no correlations were conducted in lab environments without uncertainty. Even the anomalies to those general rules fit. For example, the field sites where agreeableness did not predict performance were those with fairly straightforward answers where a correct answer is relatively easy to verify once it is put forward (e.g., the profitability of pizza franchisees (Hofmann and Jones, 2005)), and laboratory studies where agreeableness did predict performance were ones using judgment-based tasks with a strong degree of uncertainty in measuring performance (e.g., student groups work on research projects (Van Vianen and De Dreu, 2001)). Finally, our findings from this study are consistent with Deming (2017), which suggest that the labor market increasingly rewards social skills.

## Analysis

While the results of the model show new findings that are corroborated by analysis of real-world data, one key advantage of using ABMs is the fact that their behavior can be analyzed in more detail in order to understand *why* the results are as reported. In this section, we examine the effects of the GA on the ABM and how the different components of the ABM then cause the results.

Figure 9 shows the change of average Big-Five traits in each team as evolution progresses, clearly showing convergence toward different values for each trait for each experiment. For example, for best performing teams

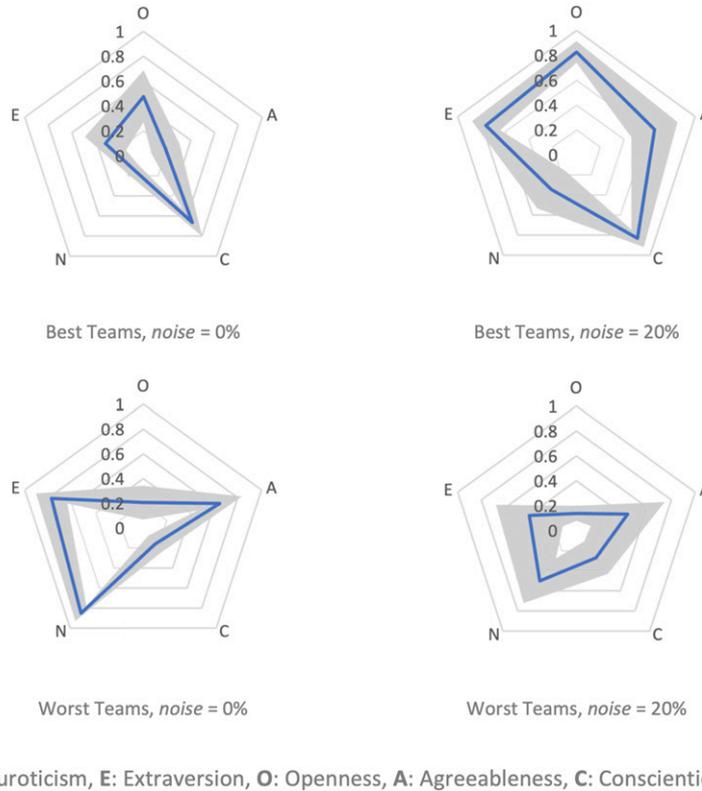

Best Teams, *noise* = 0%

Best Teams, *noise* = 20%

Worst Teams, *noise* = 0%

Worst Teams, *noise* = 20%

**N**: Neuroticism, **E**: Extraversion, **O**: Openness, **A**: Agreeableness, **C**: Conscientiousness

**Figure 6.** Average team Big-Five generated by model. Grey regions: mean ± SD. (*t*-test results show *p* = .000 comparing all traits and noise levels, see Supplemental Materials Table 2).

tackling noisy problems, neuroticism is generally lower with the other traits higher. But this is not the same for problems with no noise. The worst teams tackling problems with less noise generally have low openness and low conscientiousness and higher values for the other traits. Looking at the more constrained model, the sample population (Figure 10) displays the same trend as the general population (Figure 9) except for conscientiousness, which is relatively high regardless of best or worst teams, with or without uncertainty. This is because the individuals modelled on the dataset had a higher-than-average level of conscientiousness (replicating the fact that the MBA students in the dataset were more conscientious than the average population).

Figure 11 shows representative runs of each agent's path as it navigates the solution space to find the optimal solution in its team of four. Noise affects the path of agents, making them slower to converge. Best teams reliably converge on the optimal solution (0,0). Worst teams converge on an incorrect solution, or struggle to converge at all in the allocated time.

In the example runs, it is possible to see the effects of each personality: the best teams have generally lower average N values (neuroticism) than the worst teams. Higher average N values correspond to impulsiveness and volatility, simulated by accelerating at a random speed toward a random direction—shown in the chaotic and jagged paths for the worst compared to the smoother, cleaner paths for the best teams. As can be seen in Figure 9, the need for lower average N values was reduced with higher noise (since neuroticism adds a type of noise, for noisy problems you effectively need less neuroticism to achieve the same behavior).

The best teams tend to have higher average O values (openness) than the worst teams, corresponding to greater intellectual curiosity, simulated by having a larger area of exploration. This makes such agents more likely to find better solutions and less likely to be trapped in local optima (which might arise from the added noise), as can be seen from the better convergence to the optima for the best teams and the very poor convergence to a non-optimal for the worst teams with noise (see Figure 11).

Again, in general the better teams have higher average C values (conscientiousness) corresponding to lower spontaneity, more reliability, and less procrastination, which is simulated by less random disruption to the acceleration vector and less likely to stop moving. In Figure 11, the paths of the agents illustrate the difference, with only the worst teams showing the random, staccato movement toward random directions that low values for this trait cause.

**Table 5.** Results of OLS regression analysis.

| Predictors | DV: Team performance (tasks without uncertainty) | | | DV: Team performance (tasks with uncertainty) | | |
|---|---|---|---|---|---|---|
| | Model 1 | Model 2 | Model 3 | Model 4 | Model 5 | Model 6 |
| Intercept | −.04 (.18) | −.49 (.51) | −3.11 (2.17) | −.54 (.14)** | −.80 (.38)* | −4.82 (1.64)** |
| Agreeableness | .02 (.10) | .05 (.11) | .04 (.11) | .31 (.08)** | .31 (.08)** | .31 (.08)** |
| Conscientiousness | | .17 (.08)* | .18 (.08)* | | .08 (.06) | .06 (.06) |
| Extraversion | | .04 (.11) | .05 (.11) | | −.01 (.08) | .01 (.08) |
| Neuroticism | | .10 (.10) | .11 (.10) | | .03 (.08) | .04 (.08) |
| Openness | | −.17 (.11) | −.19 (.11) | | −.00 (.08) | .00 (.08) |
| GMAT | | | .00 (.00) | | | .01 (.00)* |
| Work experience | | | .01 (.04) | | | .01 (.03) |
| Team size | | | .13 (.05)** | | | .05 (.04) |
| $R^2$ | .00 | .01 | .03 | .03** | .03** | .04** |

Note. N = 593 teams. Table entries represent unstandardized parameter estimates with standard errors in parentheses.
* significant at $p < .05$, two-tailed; ** significant at $p < .01$, two-tailed.

**Table 6.** Results of binary logistic regression analysis.

| Predictors | Tasks without uncertainty[a] | | | Tasks with uncertainty[b] | | |
|---|---|---|---|---|---|---|
| | Exp(B) | p | 95% CI | Exp(B) | p | 95% CI |
| Intercept | .000 | .591 | | .000 | .014 | |
| Agreeableness | .643 | .584 | [.132, 3.128] | 9.831 | .005 | [1.994, 48.457] |
| Conscientiousness | 2.251 | .198 | [.655, 7.736] | 1.207 | .751 | [.378, 3.854] |
| Extraversion | 1.084 | .926 | [.195, 6.033] | 1.081 | .917 | [.249, 4.693] |
| Neuroticism | 1.243 | .781 | [.267, 5.797] | 1.090 | .909 | [.248, 4.791] |
| Openness | .118 | .007 | [.025, .558] | .696 | .630 | [.159, 3.042] |
| GMAT | 1.005 | .812 | [.964, 1.049] | 1.043 | .042 | [1.002, 1.086] |
| Work experience | 1.291 | .426 | [.689, 2.420] | 1.636 | .142 | [.848, 3.154] |
| Team size | 2.895 | .007 | [1.345, 6.232] | 1.891 | .070 | [.949, 3.767] |

[a]Binary variable of team performance where worst performing team is coded as 0 and best performing team is coded as 1, adj. $R^2$ = .176.
[b]Binary variable of team performance where worst performing team is coded as 0 and best performing team is coded as 1, adj. $R^2$ = .192.

In the model, extraversion (sociability and assertiveness) is simulated by agents either being influenced more by their neighbors' positions (especially the extroverted ones) for high average C values, or by being influenced more by their own experience for low average C values. Our results (Figures 11 and 9) show that the best teams tend to have higher average extraversion for noisy problems, which is likely to be because of the improved sharing of information, which helps such teams discover better solutions. Conversely, the worst teams tend to have lower average extraversion for noisy problems for the same reason—poor communication helps make them even worse. Interestingly, the best team had low extraversion for teams with no noise—good communication is a distraction when the path to the solution is already clear. Likewise, the worst teams had high extraversion for teams with no noise. Here, good communication helps spread misinformation caused by chaotic behavior better, and makes them worse overall.

Finally, looking further into the novel contribution to the field of psychology from this work, agreeableness (social harmony and willing to compromise) is simulated by having its direction strongly influenced by its neighbors heading (more for more assertive extroverted agents) for higher average A values (agreeableness). The best teams have low average A values for problems without noise, and high average A values for problems with noise. In this problem, the strong influence of neighbors is only helpful when your neighbors have found the solution, so for zero noise it is simply a distraction which slows convergence, but for noisy problems it becomes helpful to encourage all agents to

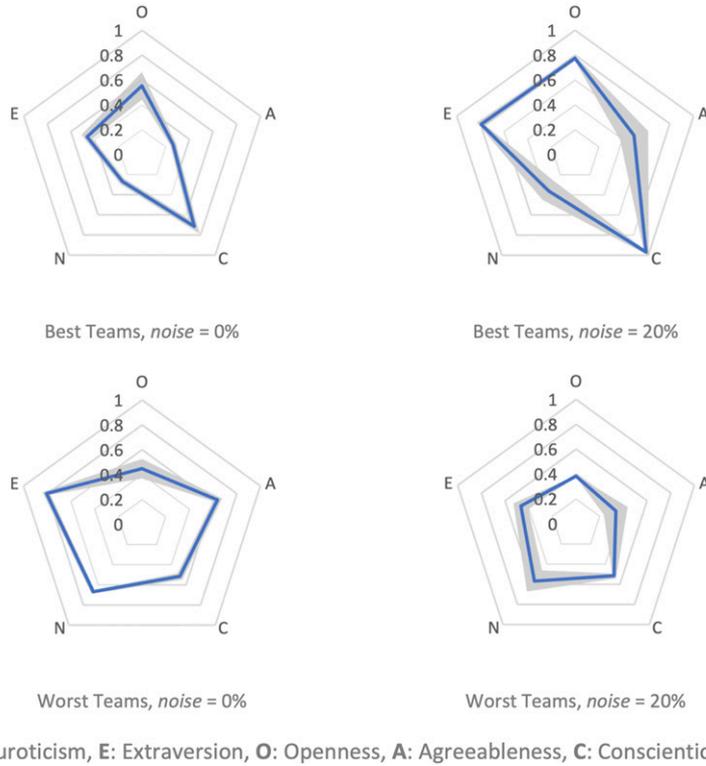

**N**: Neuroticism, **E**: Extraversion, **O**: Openness, **A**: Agreeableness, **C**: Conscientiousness

**Figure 7.** Average team Big-Five generated by model for sample population. Grey regions: mean ± SD (*t*-test results show *p* = .000 comparing all traits and noise levels, except for conscientiousness in worst teams, see Supplemental Materials Table 4).

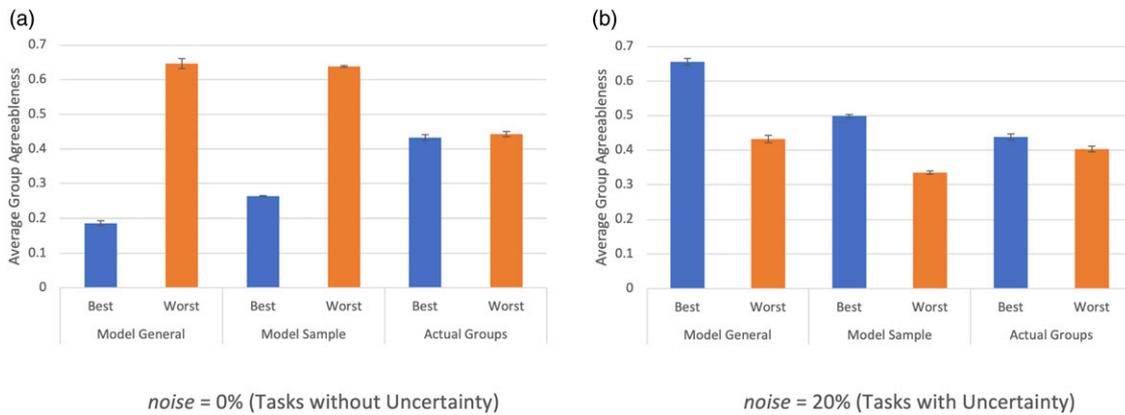

**Figure 8.** Average group agreeableness for model general population, model sample population, and actual groups, best and worst teams, tasks and without uncertainty. Error bars: mean ± SE.

converge to the solution. Conversely, the worst teams have high average A values for problems without noise and low for problems with noise for the same reason—ignoring your neighbors when tackling a noisy problem is a good way to converge on a bad solution, as can be seen in Figure 11.

Figures 9 and 10 show that there is more complexity here still to be unraveled as there are specific values for each trait that the GA converges to—at no point does it find that a trait should be eliminated or made 100% dominant in a team.

The complex interplay of the five personality traits must be carefully balanced in order to achieve a team capable of producing a desirable result. From the perspective of psychology, this points at the complexity of putting teams together with individuals having appropriate complementary traits. From the perspective of Evolutionary Computation, it shows the benefits of using a GA to discover such nuances—hand-designing an ABM would never reveal these findings.

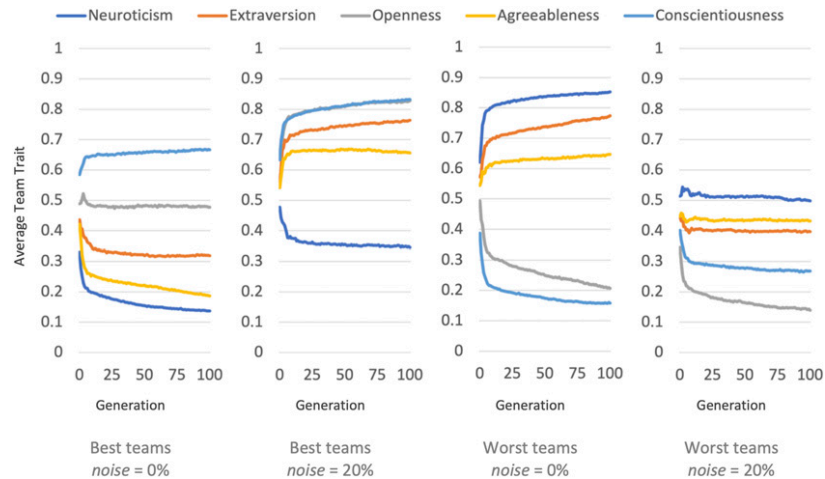

**Figure 9.** Average team Big-Five traits over generations (model general population). Plots start at first generation.

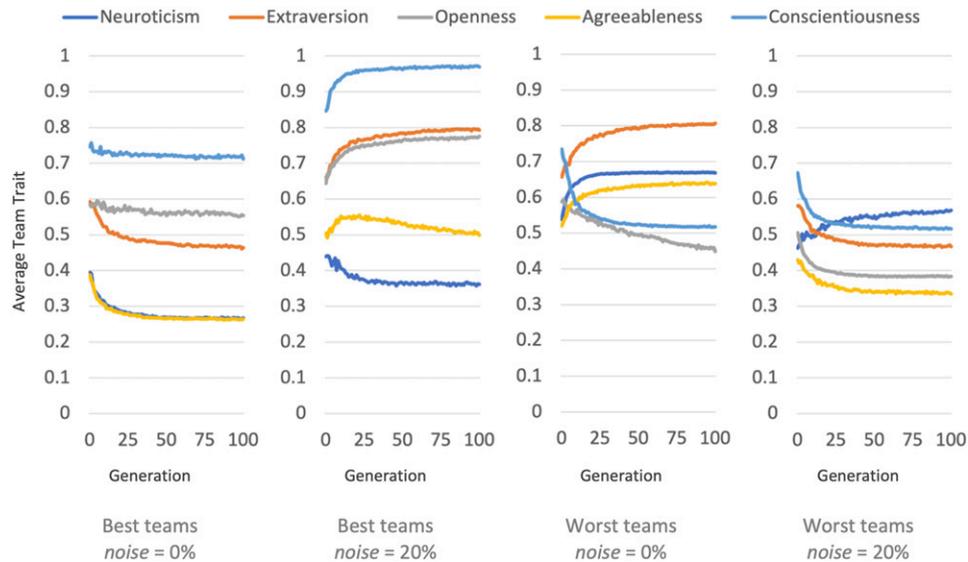

**Figure 10.** Average team Big-Five traits over generations (model sample population). Plots start at first generation.

## Limitations and future work

Although our dataset is one of the largest teamwork datasets, extensive in terms of duration (10 years) and number of teams (N = 593), it is still one dataset and from one organization. As such, there must be some caution in generalizing the results. Future work could help to establish the collection of other major teamwork datasets over many years to enable this and other models to be evaluated with multiple datasets.

Our work has shown the effects of personality on teamwork through modelling and analysis of real-world data. Personality traits may be affected by context or environment, limiting the applicability of our findings.

Nevertheless, research suggests that there is a strong genetic influence on all traits (for Neuroticism, Extraversion, Openness, Agreeableness, and Conscientiousness, broad genetic influence was estimated as 41%, 53%, 61%, 41%, and 44%, respectively), with most variance in traits caused by differing environmental influences (Jang et al., 1996); here, we assume shared environments as our team members work closely together to achieve their shared goal.

We use a GA to optimize the ABM because of its ability to handle noisy and discontinuous problems. Despite the suitability of GAs for this task, other optimizers could also be used; however, it should be noted that the purpose of this work is not to find the optimal optimizer for this ABM, it is to use the GA as a method of investigating the ABM.

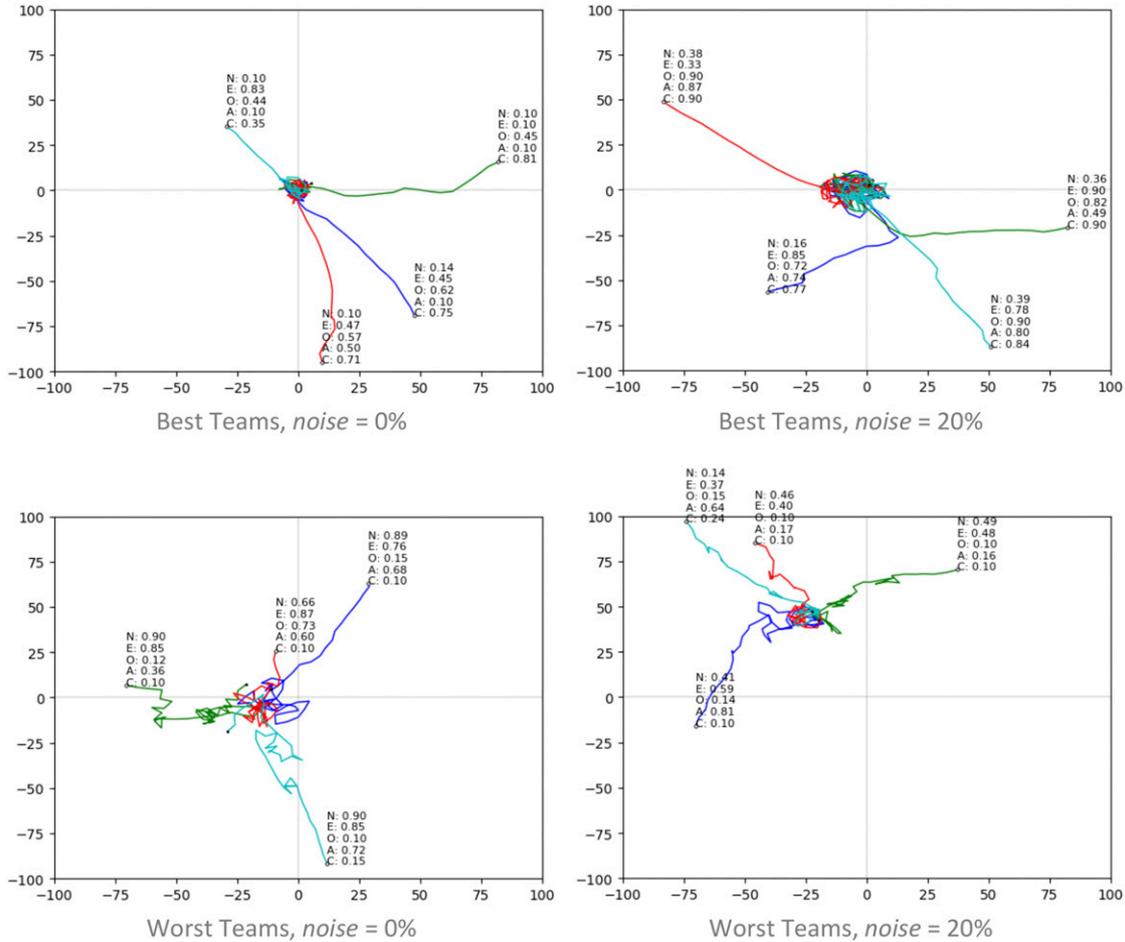

**Figure 11.** Representative runs showing each agent's path as they navigate the solution space to find the optimal solution. Black circle indicates their position at *t* = 0, and black dot indicates their position at *t* = 100. The maximum is located at the center of each image.

In this paper, a simple objective function is used because we wanted to limit the number of variables being tested and make the investigation tractable, that is, we study problem noise rather than problem difficulty in this work. There is much scope to investigate different types of group tasks and map them to different objective functions. This model is just a promising beginning: there is also an opportunity to use it to answer a variety of questions that are relevant, for example, fix some agents' personalities and find optimal compositions for the rest (i.e., supplementing an existing team with new team members for specific problems).

The model is specific for interaction of personalities for teamwork. Future work could look at advancing agent-based modelling so that the model can become more generic and applicable into other types of interaction without needing to rebuild the model. Also, there are many aspects that can affect teamwork, in addition to personalities of individual members. How best to add new elements, while still verifying the original components remain valid? The broader picture for the work is to build a simulated laboratory where we can answer questions and test hypothesis related to human behavior.

## Conclusion

Our study demonstrates the usefulness of combining evolutionary computation with ABMs for predicting the effects of personality on teamwork. By using a GA to examine the extremes of the ABM, we were able to discover the dependencies of this complex system. For noisy problems, extraversion, openness, and conscientiousness consistently positively predict team performance, while neuroticism consistently negatively predicts performance. The most surprising and novel results are related to the role of agreeableness. For tasks with noise, the best teams have higher agreeableness than worst teams, and worst teams in non-noisy tasks have lower agreeableness than worst teams in noisy tasks. These results were corroborated using a dataset that was not used for model design or validation, comprising data from one of the largest datasets on team performance published to date, comprising 3698 individuals in 593 teams collected over a 10-year period on more than 5000 group tasks with and without uncertainty. For the first time, task uncertainty (modelled as noise in this work) has been identified as a moderator of the dependency between team performance and agreeableness. This is significant because uncertainty is pervasive around the world. Technology change appears exponential, job security is low, political systems are unstable, and environmental change is accelerating. Our work implies that agreeableness may be important to facilitate teamwork and organizational performance in this new world.

Finally, this work has made use of evolutionary computation to explore the limits of the ABM for the purposes of scientific investigation. The approach provides a new methodology for the scientific investigation of teamwork, making new predictions, improving our understanding of human behaviors, and even improving team performance for organizations.

## Declaration of conflicting interests

The author(s) declared no potential conflicts of interest with respect to the research, authorship, and/or publication of this article.

## Funding

The author(s) received no financial support for the research, authorship, and/or publication of this article.

## Data Availability Statement

This article earned Open Materials badges through the study materials that can be accessed at http://www.cs.ucl.ac.uk/staff/S.Lim/team_personality_performance/.

## ORCID iD

Soo Ling Lim  https://orcid.org/0000-0003-1325-6542

## Supplemental Material

Supplemental material for this article is available online.

## Notes

1. Steiner (1972) classified tasks based on how individual contributions of members of a group are combined. In disjunctive tasks, team performance is the performance achieved by the best performing individual (e.g., a research team proving a theorem, a company searching for product market fit). Other types of tasks include conjunctive tasks where team performance is the performance achieved by the worst performing individual (e.g., a factory assembly line) and additive tasks, where team performance is the total performance of the group (e.g., a relay race or tug of war). In this work, we focus on disjunctive tasks; our previous work (Lim and Bentley, 2019a) investigated various task types.

2. http://www.cs.ucl.ac.uk/staff/S.Lim/team_personality_performance/.
3. Team personality and performance data are available at http://www.cs.ucl.ac.uk/staff/S.Lim/team_personality_performance/.

## References

Ahrndt S, Fähndrich J and Albayrak S (2015) Modelling of personality in agents: from psychology to implementation. *Proceedings of the Workshop on Human-Agent Interaction Design and Models (HAIDM) at Autonomous Agents and MultiAgent Systems (AAMAS)* Istanbul, Turkey: 1–16.

Almeida A and Azkune G (2018) Predicting human behaviour with recurrent neural networks. *Applied Sciences* 8(2): 305.

Arrow H, McGrath JE and Berdahl JL (2000) *Small Groups as Complex Systems: Formation, Coordination, Development, and Adaptation*. Thousand Oaks, California: Sage Publications.

Barrick MR, Stewart GL, Neubert MJ, et al. (1998) Relating member ability and personality to work-team processes and team effectiveness. *Journal of Applied Psychology* 83(3): 377–391.

Barry B and Stewart GL (1997) Composition, process, and performance in self-managed groups: the role of personality. *The Journal of Applied Psychology* 82(1): 62–78.

Bell ST (2007) Deep-level composition variables as predictors of team performance: a meta-analysis. *The Journal of Applied Psychology* 92(3): 595–615.

Bentley PJ (2009) Methods for improving simulations of biological systems: systemic computation and fractal proteins. *Journal of the Royal Society, Interface* 6(Suppl 4): S451–S466.

Bentley PJ and Lim SL (2022) From evolutionary ecosystem simulations to computational models of human behavior. *Wiley Interdisciplinary Reviews: Cognitive Science* 13(6): e1622.

Bradley BH, Baur JE, Banford CG, et al. (2013) Team players and collective performance: how agreeableness affects team performance over time. *Small Group Research* 44(6): 680–711.

Bryson JJ, Ando Y and Lehmann H (2007) Agent-based modelling as scientific method: a case study analysing primate social behaviour. *Philosophical Transactions of the Royal Society B: Biological Sciences* 362(1485): 1685–1699.

Bunn DW and Oliveira FS (2001) Agent-based simulation-an application to the new electricity trading arrangements of England and Wales. *IEEE Transactions on Evolutionary Computation* 5(5): 493–503.

Chen J-C, Lin T-L and Kuo M-H (2002) Artificial worlds modeling of human resource management systems. *IEEE Transactions on Evolutionary Computation* 6(6): 542–556.

Cordery JL, Morrison D, Wright BM, et al. (2010) The impact of autonomy and task uncertainty on team performance: a longitudinal field study. *Journal of Organizational Behavior* 31(2-3): 240–258.

Costa PT and MacCrae RR (1992) *Revised NEO Personality Inventory (NEO PI-R) and NEO Five-Factor Inventory (NEO-FFI): Professional Manual*. Odessa, Florida: Psychological Assessment Resources, Inc.

Cuthbert MO, Gleeson T, Reynolds SC, et al. (2017) Modelling the role of groundwater hydro-refugia in East African hominin evolution and dispersal. *Nature Communications* 8: 15696. DOI: 10.1038/ncomms15696

Cziko GA (1989) Unpredictability and indeterminism in human behavior: arguments and implications for educational research. *Educational Researcher* 18(3): 17–25.

Day AL and Carroll SA (2004) Using an ability-based measure of emotional intelligence to predict individual performance, group performance, and group citizenship behaviours. *Personality and Individual Differences* 36(6): 1443–1458.

Deming DJ (2017) The growing importance of social skills in the labor market. *The Quarterly Journal of Economics* 132(4): 1593–1640.

DeYoung CG, Quilty LC and Peterson JB (2007) Between facets and domains: 10 aspects of the Big Five. *Journal of Personality and Social Psychology* 93(5): 880–896.

Epstein JM and Axtell R (1996) *Growing Artificial Societies: Social Science from the Bottom up*. Washington D.C.: Brookings Institution Press.

Fabi V, Andersen RV, Corgnati SP, et al. (2013) A methodology for modelling energy-related human behaviour: application to window opening behaviour in residential buildings. *Building Simulation*. Berlin Heidelberg: Springer, 6.

Farmer JD and Foley D (2009) The economy needs agent-based modelling. *Nature* 460(7256): 685–686.

Fleeson W and Gallagher P (2009) The implications of Big Five standing for the distribution of trait manifestation in behavior: fifteen experience-sampling studies and a meta-analysis. *Journal of Personality and Social Psychology* 97(6): 1097–1114.

Goldberg LR (1990) An alternative" description of personality": the big-five factor structure. *Journal of Personality and Social Psychology* 59(6): 1216.

Gonzalez MC, Hidalgo CA and Barabasi A-L (2008) Understanding individual human mobility patterns. *Nature* 453(7196): 779–782.

Guo S, Lim SL and Bentley PJ (2020) Teams frightened of failure fail more: modelling reward sensitivity in teamwork. *Proceedings of the 2020 IEEE Symposium Series on Computational Intelligence (SSCI)*, Canberra, ACT, Australia: 109–116.

Hofmann DA and Jones LM (2005) Leadership, collective personality, and performance. *The Journal of Applied Psychology* 90(3): 509.

Holland JH (1992) *Adaptation in Natural and Artificial Systems*. 2nd ed. Cambridge, MA: MIT Press.

Izumi K and Ueda K (2001) Phase transition in a foreign exchange market-analysis based on an artificial market approach. *IEEE Transactions on Evolutionary Computation* 5(5): 456–470.

Jang KL, Livesley WJ and Vemon PA (1996) Heritability of the big five personality dimensions and their facets: a twin study. *Journal of Personality* 64(3): 577–592.

Kennedy J, Eberhart RC and Shi Y (2001) *Swarm Intelligence*. San Francisco, California: Morgan Kaufmann.

Kim BS, Kang BG, Choi SH, et al. (2017) Data modeling versus simulation modeling in the big data era: case study of a greenhouse control system. *Simulation* 93(7): 579–594.

Komarraju M and Karau SJ (2005) The relationship between the big five personality traits and academic motivation. *Personality and Individual Differences* 39(3): 557–567.

Laughlin PR and Ellis AL (1986) Demonstrability and social combination processes on mathematical intellective tasks. *Journal of Experimental Social Psychology* 22(3): 177–189.

Lim SL and Bentley PJ (2018) Coping with uncertainty: modelling personality when collaborating on noisy problems. *Proceedings of the 2018 Conference on Artificial Life*, Tokyo, Japan: 566–573.

Lim SL and Bentley PJ (2019a) All in good team: optimising team personalities for different dynamic problems and task types. *Proceedings of the 2019 Conference on Artificial Life*, Newcastle, United Kingdom: 153–160.

Lim SL and Bentley PJ (2019b) Diversity improves teamwork: optimising teams using a genetic algorithm. *Proceedings of the IEEE Congress on Evolutionary Computation*, Wellington, New Zealand: 2848–2855.

Lim SL, Bentley PJ and Ishikawa F (2015) The effects of developer dynamics on fitness in an evolutionary ecosystem model of the App Store. *IEEE Transactions on Evolutionary Computation* 20(4): 529–545.

Marchau VA, Walker WE, Bloemen PJ, et al. (2019) *Decision Making under Deep Uncertainty: From Theory to Practice*. Cham, Switzerland: Springer Nature.

Minar N, Burkhart R, Langton C, et al. (1996) *The Swarm Simulation System: A Toolkit for Building Multi-Agent Simulations*. Santa Fe, New Mexico: Santa Fe Institute.

Mohammed S and Angell LC (2003) Personality heterogeneity in teams: which differences make a difference for team performance? *Small Group Research* 34(6): 651–677.

Morrall D (2003) Ecological applications of genetic algorithms. *Ecological Informatics*. Berlin Heidelberg: Springer-Verlag, 35–48.

Moynihan LM and Peterson RS (2001) A contingent configuration approach to understanding the role of personality in organizational groups. *Research in Organizational Behavior* 23: 327–378.

Muthukrishna M and Henrich J (2019) A problem in theory. *Nature Human Behaviour* 3(3): 221–229.

Neuman GA, Wagner SH and Christiansen ND (1999) The relationship between work-team personality composition and the job performance of teams. *Group and Organization Management* 24(1): 28–45.

Neuman GA and Wright J (1999) Team effectiveness: beyond skills and cognitive ability. *The Journal of Applied Psychology* 84(3): 376–389.

Nicolaisen J, Petrov V and Tesfatsion L (2001) Market power and efficiency in a computational electricity market with discriminatory double-auction pricing. *IEEE Transactions on Evolutionary Computation* 5(5): 504–523.

Nikolic I and Dijkema GP (2010) On the development of agent-based models for infrastructure evolution. *International Journal of Critical Infrastructures* 6(2): 148–167.

Palmer R, Arthur WB, Holland JH, et al. (1994) Artificial economic life: a simple model of a stockmarket. *Physica D: Nonlinear Phenomena* 75(1): 264–274.

Peeters MA, Van Tuijl HF, Rutte CG, et al. (2006) Personality and team performance: a meta-analysis. *European Journal of Personality* 20(5): 377–396.

Peterson RS, Smith DB, Martorana PV, et al. (2003) The impact of chief executive officer personality on top management team dynamics: one mechanism by which leadership affects organizational performance. *The Journal of Applied Psychology* 88(5): 795–808.

Resnick P and Varian HR (1997) Recommender systems. *Communications of the ACM* 40(3): 56–58.

Sagl G, Resch B, Hawelka B, et al. (2012) *From social sensor data to collective human behaviour patterns: analysing and visualising spatio-temporal dynamics in urban environments. Proceedings of the GI-Forum*, Berlin: Herbert Wichmann Verlag: 54–63.

Salvit J and Sklar E (2012) Modulating agent behavior using human personality type. *Proceedings of the Workshop on Human-Agent Interaction Design and Models (HAIDM) at Autonomous Agents and MultiAgent Systems (AAMAS)*, Valencia, Spain: 145–160

Sampaio A, Soares JM, Coutinho J, et al. (2014) The Big Five default brain: functional evidence. *Brain Structure and Function* 219(6): 1913–1922.

Smith VL (1991) *Papers in Experimental Economics*. Cambridge: Cambridge University Press.

Steiner ID (1972) *Group Process and Productivity*. New York: Academic Press.

Takadama K, Terano T and Shimohara K (2001) Nongovernance rather than governance in a multiagent economic society. *IEEE Transactions on Evolutionary Computation* 5(5): 535–545.

Tassier T and Menczer F (2001) Emerging small-world referral networks in evolutionary labor markets. *IEEE Transactions on Evolutionary Computation* 5(5): 482–492.

Van Vianen AE and De Dreu CK (2001) Personality in teams: its relationship to social cohesion, task cohesion, and team performance. *European Journal of Work and Organizational Psychology* 10(2): 97–120.

Whigham P and Fogel G (2006) Ecological applications of evolutionary computation. *Ecological Informatics*. Berlin Heidelberg: Springer-Verlag, 85–107.

Wood RE (1986) Task complexity: definition of the construct. *Organizational Behavior and Human Decision Processes* 37(1): 60–82.

Woolley AW, Chabris CF, Pentland A, et al. (2010) Evidence for a collective intelligence factor in the performance of human groups. *Science* 330(6004): 686–688.